\documentclass{article}

\usepackage[nonatbib,final]{nips_2017}

\usepackage{amsmath,amsthm,amssymb}

\DeclareMathOperator*{\argmin}{arg\,min} 
\makeatletter
\g@addto@macro\normalsize{%
  \setlength\abovedisplayskip{1pt}
  \setlength\belowdisplayskip{1pt}
}

\allowdisplaybreaks[4]
\makeatother
\usepackage{tikz}
\usetikzlibrary{bayesnet}
\usepackage{rotating}
\usepackage{pdflscape}

\usepackage[utf8]{inputenc} 
\usepackage[T1]{fontenc}    
\usepackage{hyperref}       
\usepackage{url}            
\usepackage{booktabs}       
\usepackage{amsfonts}       
\usepackage{nicefrac}       
\usepackage{microtype}      

\title{Overpruning in Variational Bayesian Neural Networks}

\author{
  Brian L. Trippe\\
  University of Cambridge and \\
  Massachusetts Institute of Technology\\
   \texttt{btrippe@mit.edu} \\
   \And
  Richard E. Turner \\
  University of Cambridge\\
   \texttt{ret26@cam.ac.uk} \\
}

\begin{document}

\maketitle

\begin{abstract}
The motivations for using variational inference (VI) in neural networks differ significantly from those in latent variable models.  This has a counter-intuitive consequence; more expressive variational approximations can provide significantly worse predictions as compared to those with less expressive families.  In this work we make two contributions. First, we identify a cause of this performance gap, variational over-pruning.  Second, we introduce a theoretically grounded explanation for this phenomenon.  Our perspective sheds light on several related published results and provides intuition into the design of effective variational approximations of neural networks.
\end{abstract}

\section{Introduction}
\vspace{-5pt}
  Though deep neural networks have been enormously successful across a variety of prediction tasks, they often fail to accurately capture uncertainty, a characteristic which has motivated a resurgence of interest in Bayesian methods for learning neural networks.  Following initial work on using variational inference (VI) to fit neural networks \cite{Hinton1993,Barber1998}, a great deal of recent work has proposed new approaches for VI in these models \cite{Graves2011PracticalNetworks.,Blundell2015,Gal2015,Kingma2015,Louizos2016}.  However, VI remains difficult and the performance benefits as compared to other approaches for capturing uncertainty is unclear \cite{Lakshminarayanan2017}.

The motivations for using VI to capture parameter uncertainty differ significantly from those for using VI in latent variable models for which we are inherently concerned with a posterior over hidden variables.  In Bayesian approximations of neural networks, the posterior over weights and biases generally is not the object of interest; instead we are concerned with the posterior over functions.  As a result, failing to capture characteristics of the exact posterior such as multi-modality (which we know to be imparted by the many symmetries and degeneracies of NNs) is not necessarily problematic.  Empirically, methods that perform VI over parameters that we know not to even resemble the exact posterior over parameters can perform acceptably in practice, providing reasonably well calibrated uncertainty predictions on small datasets \cite{Gal2015,Blundell2015}.

The organization of this short paper is as follows.  In section \ref{sec:consequence} we document a surprising consequence of this mismatch; more expressive variational approximations can provide worse performance than less expressive ones.  Next, in section \ref{sec:pruning} we identify a cause of this performance gap, the over-pruning of hidden units.  Finally, in section \ref{sec:tradeoff} we provide a theoretical explanation for this phenomenon which clearly explains over-pruning as well as several other peculiar results in variational Bayesian neural network literature.

\vspace{-5pt}
\section{Variational Approximations of Neural Networks}\label{sec:consequence}
\vspace{-5pt}
We consider supervised learning problems in which we have a dataset $\mathcal{D}=\{x_i,y_i\}_{i=1}^N$ of observation/label pairs, $(x_i\in X, y_i \in Y)$, sampled i.i.d. from some joint distribution, $p(x,y)$, and are interested in estimating the conditional, $p(y_{\mathrm{new}}|x_{\mathrm{new}})$.  We suppose the labels are sampled from a discriminative probabilistic model parameterized by $\theta$, such that $y_i \sim p(y|x_i,\theta)$ and use $\mathcal{D}$ to learn about $\theta$ in order make predictions.

The Bayesian approach considers $\theta$ to be an unknown variable, places a prior over it (here parameterised by $\alpha$) and seek the posterior $p(\theta|\mathcal{D},\alpha) = \frac{p(\theta|\alpha)p(Y|X,\theta)}{\int_{\hat\theta} p(\hat\theta|\alpha)p(Y|X,\hat\theta) d \hat \theta}$.  In our case, $\theta$ defines the parameters of a neural network, and we make predictions by approximating the marginal over $\theta$ with a Monte Carlo estimate:
\begin{equation}
p(y_{\mathrm{new}}|x_{\mathrm{new}},\mathcal{D},\alpha) = \mathbb{E}_{p(\theta|\mathcal{D},\alpha)}[p(y_{\mathrm{new}}|x_{\mathrm{new}}, \theta)] \approx \frac{1}{M} \sum_{i=1}^M p(y_{\mathrm{new}}|x_{\mathrm{new}}, \theta_i)
\label{eqn:post_predictive}
\end{equation}

Where $M$ is the number of Monte Carlo samples of $\theta_i \sim p(\theta|\mathcal{D}, \alpha)$.  Variational inference (VI) \cite{Jordan1998} minimizes the KL-divergence between an approximation of the intractable posterior, $q(\theta)$, and $p(\theta|\mathcal{D},\alpha)$:
\begin{equation}\argmin_{q \in \mathcal{Q}} \mathrm{KL}[q||p] = \argmin_{q \in \mathcal{Q}} \mathcal{F}_{\mathrm{VFE}}(q) =  -\mathbb{E}_{ q(\theta)}\big[\mathrm{log} \  p(Y, \theta|X, \alpha)-\mathrm{log}\  q(\theta)\big]\end{equation}\label{eqn:vfe}

The choice of variational family, $\mathcal{Q}$, is important when employing variational approximations of neural networks, and a number of different variational families have been proposed \cite{Graves2011PracticalNetworks.,Gal2015,Louizos2016,Louizos2017MultiplicativeNetworks}.  We explore the performance of several variational families when applied to a single hidden layer MLP  with $50$ hidden units and tanh activations on six benchmark UCI regression datasets.  The methods include maximum likelihood inference with early stopping (ES), maximum a posteriori using  a Gaussian prior over weights (MAP), variational inference with a mean-field Gaussian approximate posterior over weights with learned variances (MF) \cite{Kingma2015}, mean-field Gaussian with fixed variances (WN)\footnote{We refer to this model as Weight Noise(WN) due to its equivalence to simply adding constant variance noise to weights.  In this sense, the model is similar to fast dropout \cite{Wang2013} and drop-connect \cite{Wan2013} .} and Gaussian approximate posterior with full rank covariance within each layer (FC).  For each of these models we tuned hyper-parameters on a held-out validation set.  Notably, this included the prior variance for MAP, WN, MF and FC, and additionally included weight and bias variances for WN.  We additionally include the performance a sampling method, hybrid Monte Carlo (HMC) \cite{Neal1995} as previously evaluated on these same datasets by \cite{Bui2016}.

\begin{figure}
\centering
\includegraphics[width=.9\textwidth]{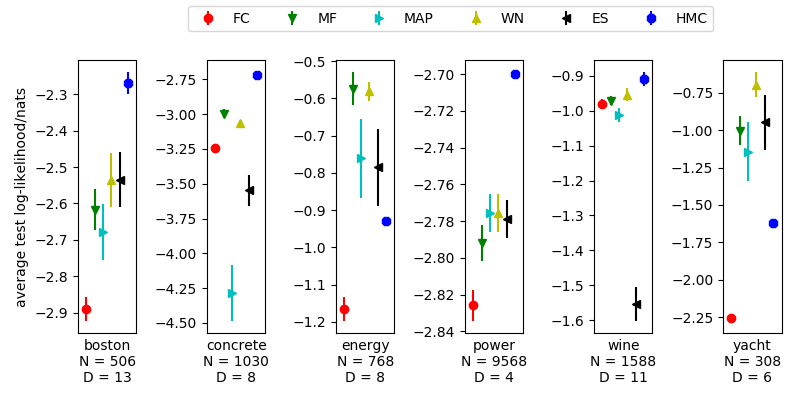} 
\caption{Comparison of performance Bayesian neural networks in terms of mean log likelihood on test sets using different variational families on six small UCI datasets.  Higher is better.  Uncertainty bars represent $\pm 1$ standard error in the mean.}
\label{fig:vi_uci_compare}
\end{figure}

\cite{Bui2016}
As we see in Figure \ref{fig:vi_uci_compare}, WN most consistently has good performance as compared to the other variational approximations tested.  MF performs well on most datasets with best performance on concrete.  ES and MAP have high variance their performance, both across tasks and across train/test splits within each task; each of these methods have the greatest performance on one dataset and the worst performance on at least one dataset.

FC performs worse than the mean-field approximation.  This is the case on all datasets other than wine quality prediction, on which the two methods have roughly equivalent performance.  This could be explained by the use of the local reparameterization trick \cite{Kingma2015} on MF and WN, which we were unable to use for FC, which used a more structured variational family.

Surprisingly, we see that in several cases that richer variational approximations perform worse than less flexible ones.  We note that the family of approximations which can be represented by WN is a strict subset of those which can be represented by MF, which itself is a strict subset of those representable by FC.  Though the more expressive families can achieve a lower variational free energy (or equivalently, a better expectation lower bound), this does not lead to better predictions\footnote{The differently chosen priors for each of the models tested precludes an informative comparison of these bounds.}.
\vspace{-5pt}
\section{Over-pruning in Mean Field Approximations}\label{sec:pruning}
\vspace{-5pt}

\begin{figure}
\centering
\includegraphics[width=1.\textwidth]{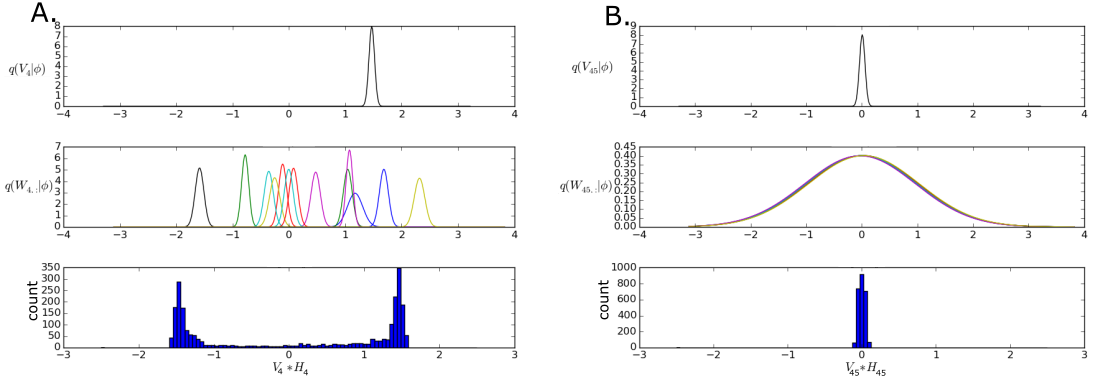} 
\caption{Representative examples of the characteristics hidden units in a neural network with tanh activation functions trained by variational inference with a factorized Gaussian approximate posterior.  A.) an active hidden unit B.) a pruned hidden unit.  Top.) the approximate posterior over the weight connecting it to the output.  Middle.) posteriors over weights connecting from the input to the hidden unit Bottom.) Histogram of $25$ sampled activations across all training data-points.   These reflect learned weights for the same network discussed in Figure \ref{fig:pruning_curves}.  Similar results for the remaining hidden units are included in the appendix (Figure \ref{fig:pruning_curves_App}).}
\label{fig:pruning_rep}
\end{figure}

To better understand why the mean field variational family performs worse than the weight noise model, we took a closer look at a posterior approximation fit to the `Boston Housing' dataset.  Watching the trajectories of the $\mathrm{KL}$ divergence, expected negative log-likelihood and negative log-likelihood under the posterior predictive distribution, we make several observations (Appendix Figure \ref{fig:pruning_curves}).  As anticipated, the estimated $\mathcal{F}_{\mathrm{VFE}}$ decreases monotonically throughout the course of optimization.  The negative log-likelihood under the posterior predictive distribution, however, shows unexpected behavior.  For both the training and test sets, this term initially decreases but then rebounds slightly, increasing and converging to larger a value than it had taken earlier.  While an increase in the negative log expected likelihood of test data is a sign of overfitting, this behavior on the training set indicates underfitting.

Looking more closely at the posterior approximation, we see that most of the hidden units have been pruned away (Figure \ref{fig:pruning_rep}).  Of the $50$ hidden units, $39$ have learned output weight confidently around zero with all incoming weights sitting precisely at the prior, and the remaining $11$ have learned outputs weights farther from zero with incoming weights that are more certain and dispersed.  We refer to this phenomenon as variational overpruning and believe it is largely responsible for the performance decay seen in figure \ref{fig:pruning_rep}.

\begin{figure}
\centering
\includegraphics[width=1.\textwidth]{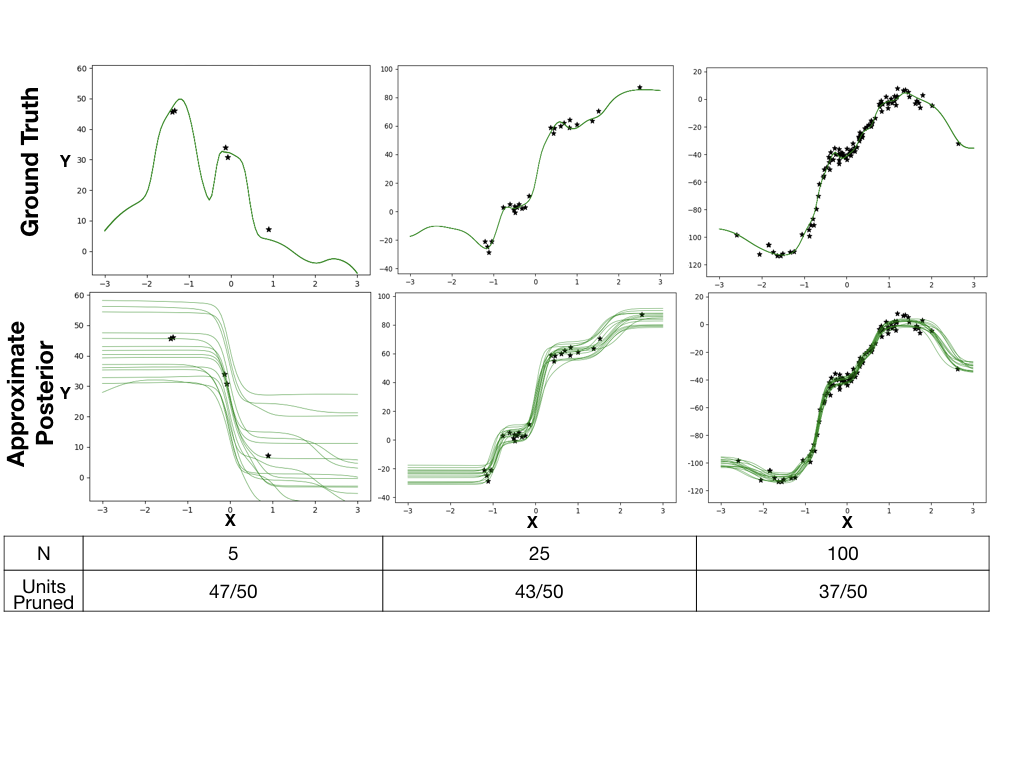} 
\vspace{-75pt}\caption{Pruning occurs in correctly specified models with good initialization and induces a biased posterior over functions \textbf{Top.)} Functions and datasets are sampled from a NN with 50 hidden units by sampling weights from the prior, mapping inputs through the function and adding observation noise.  Three datasets are simulated with N$=5$, $25$ and $100$ (from left to right) \textbf{Bottom.)} Pruning occurs robustly and functions sampled from the learned approximate posteriors are biased, with steps at the inflection points of the tanh activation function of the unpruned hidden units.  Less pruning occurs for larger N.  Posterior means were initialized to the `true' weights with low uncertainty.}
\label{fig:correct_model}
\end{figure}

To investigate the possibility that the observed pruning was an artifact the optimisation \footnote{This possibility initially seemed plausible given that the objective is non convex, and optimisation relies of noisy gradient estimates}, we performed an experiment in a more simplified setting (Figure \ref{fig:correct_model}).  We simulated data from a neural network by sampling weights and biases from a Gaussian prior with Gaussian observation noise, and performed mean field inference initializing the variational approximation with mean equal to the true parameters and a small initial variance.  This construction allows us to observe the behavior of a correctly specified model.

As the uncertainty in the approximate posterior increases from its small initialization many hidden units are pruned away: the mean of the output weights tends to zero and the variance to a very small value.  As a result, functions within the support of the approximate posterior consist of a small number of steps, each corresponding to the tanh nonlinearity of an un-pruned hidden unit. This is manifestly wrong from a Bayesian perspective: when the number of data are small, we should be very uncertain about most of the output weights, rather than confident that they are zero. Overpruning limits the expressiveness of these models and leads to under-fitting, in both a toy example and in real regression problems.  In the next section, we explain how this phenomenon is a manifestation of a more general problem with variational methods\cite{Turner1998}.
\vspace{-5pt}
\section{Tightness of the Variational Bound Explains Overpruning}\label{sec:tradeoff}
\vspace{-5pt}
In this section, we propose theoretically grounded explanation for overpruning.  We can gain insight into the performance of the mean-field approximation by decomposing $\mathcal{F}_{\mathrm{VFE}}(q)$ into the expected log likelihood and a complexity penalty for each layer.  This decomposition provides a clear explanation for the source of variational overpruning.

To be concrete, consider an MLP with a single hidden layer of $H$ hidden units with activations denoted as $\mathbf{h}=(h_1, h_2, \dots, h_N)$, defined as a function of weight matrix, $W$, and input, $x$, as $\mathbf{h} = \mathrm{tanh}(W\cdot x)$, and a single output defined as the dot product of a second weight matrix, $V$, with $\mathbf{h}$ such that $f_\theta(x) = V \cdot \mathrm{tanh}(W\cdot x)$\footnote{We neglect biases for simplicity.  With biases, the argument is identical but notationally more complex.}.  As such, we have parameters $\theta= (W,V)$, and can write the variational objective as:
\begin{equation}
\begin{aligned}
\mathcal{F}_{\mathrm{VFE}} &=  -\mathbb{E}_{ q(\theta)} \big[\mathrm{log} \  p(Y|X, \theta)\big]+ \mathrm{D_{KL}}\big(q(\theta)||p(\theta|\alpha)\big) \\
&=  -\mathbb{E}_{ q(\theta)} \big[\mathrm{log} \  p(Y|X, \theta)\big]+ \sum_{j=1}^H \mathrm{D_{KL}}\big(q(v_j)||p(v_j|\alpha)\big) + \sum_{i=1}^D \mathrm{D_{KL}}\big(q(w_{j,i})||p(w_{j,i}|\alpha)\big) \\
\end{aligned}
\end{equation}
This presentation of $\mathcal{F}_{\mathrm{VFE}}(q)$ as the sum of the expected log-likelihood and the complexity penalty (the KL-divergence of $q(\theta)$ from the prior, $p(\theta|\alpha)$) defines a trade-off between modeling the complexity of the data and retaining the simplicity of the prior \cite{Blundell2015}. The pruning of hidden units as in figure \ref{fig:pruning_rep}B reduces the tension of this trade-off.   As we see in figure \ref{fig:pruning_rep}B, when the approximate posterior over a hidden-to-output weight, $v_j$, is centered on $0$ with low uncertainty, the corresponding hidden unit, $h_j$, no longer impacts the output.  In turn, the incoming weights to $h_j$, $w_{j,:}$ no longer have an impact on the expected log-likelihood.  As a result:

$$
p(w_{j,i}|v_{j}=0,\mathcal{D},\alpha)=\frac{p(w_{j,i}|v_{j}=0,\alpha)p(\mathcal{D}|v_{j}=0,\alpha)}{p(\mathcal{D}|v_{j}=0,\alpha)}=p(w_{j,i}|\alpha)
$$

In this way, learning variational approximations $q(v_j)  \approx \delta(0)$, establishes conditional independence between each $w_{j,i}$ and the data, and $p(w_{j,i}| v_{j}=0, \mathcal{D},\alpha)$ collapses to its prior. In a network with a single output and multiple inputs, incoming weights far outnumber output weights and pruning provides a mechanism for reducing the complexity penalty without incurring a large penalty for increasing the variance in predictions (as would occur if the output weights were uncertain as well). This mechanism reduces the variational free energy by bringing the exact posterior closer to the prior rather than by explaining the data.  This is reminiscent of the known property of variational maximizations that the tightness of the variational bound induces biases in parameter estimates \cite{Turner1998}.

We argue that variational over-pruning is a common pathology of variational approximations to neural networks in which variances are learned, and believe the effect of pruning is compatible with several surprising documented observations.  For example, Blundell et al.~\cite{Blundell2015} showed that up to $98\%$ of the weights of a network trained on MNIST could be pruned with an accuracy decay of only $0.15\%$ (from $1.24\%$ error to $1.39\%$) \footnote{It is worth noting, however, that the performance of their approach did not surpass that of drop-connect which can be viewed as using a less expressive variational family\cite{Wan2013}}.  A separate result which may be related to variational over-pruning was published by Molchanov et al.~\cite{Molchanov2017}, who showed that, when optimizing the dropout probabilities of networks using variational dropout, many drop probabilities drifted to $1$.  They referred to this property as inducing sparsity, but the results of this paper suggest that it might be due to overpruning instead.  The overpruning may also explain why variational Bayesian neural networks can be compressed to such a high degree without significant loss in performance  \cite{Louizos2017BayesianLearning}.  However, generally speaking, in our eyes, this behaviour is a shortcoming of  variational inference rather than a feature.

In contrast to MF, WN does not have learned variances and is therefore unable to prune hidden units.  This encourages learning of parameters which define smoother functions and does not underfit, thus explaining the observed performance gap.
\vspace{-5pt}
\section{Conclusion}
\vspace{-5pt}
We have demonstrated a surprising property of variational approximations to neural networks; expressive approximations can provide worse performance than more constrained approximations.  We identified variational overpruning as an explanation for this phenomenon and provided a theoretical explanation for why it occurs. Despite much recent work on improving variational approximations to neural networks, theoretical justification for the use of one family over another is largely absent and it is often unclear how to choose variational families.  We hope our perspective provides a grounding for the selection of variational approximations.
\vspace{-5pt}
\section{Acknowledgments}
The authors would like to thank Thang Bui, Yingzhen Li, and Cuong Nyguyen for insightful comments and discussion and the reviewers for constructive feedback.
\vspace{-5pt}

\bibliographystyle{plain} 
\bibliography{Mendeley.bib} 

\medskip

\appendix
\section{Experimental Details}
We use a single hidden layer MLP with $50$ hidden units and tanh activations.  Our priors and approximate posteriors are diagonal Gaussian.  We use a unit normal prior on both weights and biases.  We initialize the posterior uncertainties to be $10^{-4}$.  We train using Adam with parameters $\beta^1=0.9$ and $\beta^2=0.99$ with a learning rate of $0.005$.  We ran our optimization for $5000$ iterations of batch optimization, except for ES, for which we ran for $2000$ epochs (we did not use a validation set, but instead optimized the learning rate on a held out dataset).  

We use these the same $20$ train-test splits as these previous methods.  We optimize hyper-parameters on one of the train/test splits by Bayesian optimization using Spearmint \footnote{\href{https://github.com/HIPS/Spearmint}{Spearmint} is openly available at \href{https://github.com/HIPS/Spearmint}{https://github.com/HIPS/Spearmint}}\cite{Snoek2012PracticalAlgorithms}).  For ES, we optimize the learning rate.  For all other models we optimized the prior standard deviation.  For WN, we additionally optimize the standard deviation of the approximate posterior by Bayesian optimization.  Following \cite{Blundell2015}, we initialize the variances on weights to be very small ( $\sigma_{\mathrm{init}}=10^{-4}$), a trick which empirically seems to provide better results.

\begin{table}
\caption{Held-out mean log-likelihood in nats $\pm 1 SEM$  for several variational methods for fitting a $50$ hidden unit neural network benchmark regression datasets.  Higher is better.}
\centering
\label{table:uci_results_vi_cpy}
\fontsize{7.45}{9.325}\selectfont

{\begin{tabular}{|p{0.8cm}|p{0.4cm}|p{0.2cm}|p{1.2cm}|p{1.2cm}|p{1.2cm}|p{1.2cm}|p{1.2cm}|p{1.2cm}|p{1.2cm}|p{1.2cm}|}

\iftrue
\hline
\textbf{Dataset} & N & D & \textbf{HMC} & \textbf{SGLD}  & \textbf{ES} & \textbf{FC} & \textbf{MAP} & \textbf{MF} & \textbf{WN}\\
\hline
boston & 506 & 13 & \textbf{-2.27}$\pm$\textbf{0.03} & -2.40$\pm$0.05 & -2.53$\pm$0.08 & -2.89$\pm$0.03 & -2.68$\pm$0.08 & -2.62$\pm$0.06 & -2.54$\pm$0.07 \\
concrete & 1030 & 8 & \textbf{-2.72}$\pm$\textbf{0.02} & -3.08$\pm$0.03 & -3.55$\pm$0.11 & -3.24$\pm$0.01 & -4.29$\pm$0.20 & -3.00$\pm$0.03 & -3.07$\pm$0.03 \\
energy & 768 & 8 & -0.93$\pm$0.01 & -2.39$\pm$0.01 & -0.62$\pm$0.03 & -1.16$\pm$0.03 & -0.76$\pm$0.10 & \textbf{-0.57}$\pm$\textbf{0.04} & -0.58$\pm$0.03 \\
power & 9568 & 4 & -2.70$\pm$0.00 & \textbf{-2.67}$\pm$\textbf{0.00} & -2.78$\pm$0.01 & -2.83$\pm$0.01 & -2.78$\pm$0.01 & -2.79$\pm$0.01 & -2.78$\pm$0.01 \\
wine & 1588 & 11 & -0.91$\pm$0.02 & \textbf{-0.41}$\pm$\textbf{0.01} & -1.55$\pm$0.05 & -0.98$\pm$0.01 & -1.01$\pm$0.02 & -0.97$\pm$0.01 & -0.95$\pm$0.02 \\
yacht & 308 & 6 & -1.62$\pm$0.01 & -2.90$\pm$0.02 &  -0.95$\pm$0.18 & -2.26$\pm$0.02 & -1.14$\pm$0.20 & -1.00$\pm$0.10 & \textbf{-0.70}$\pm$\textbf{0.08} \\
\hline
\fi

\end{tabular}}
\end{table}

\begin{figure}
\centering
\includegraphics[width=1.\textwidth]{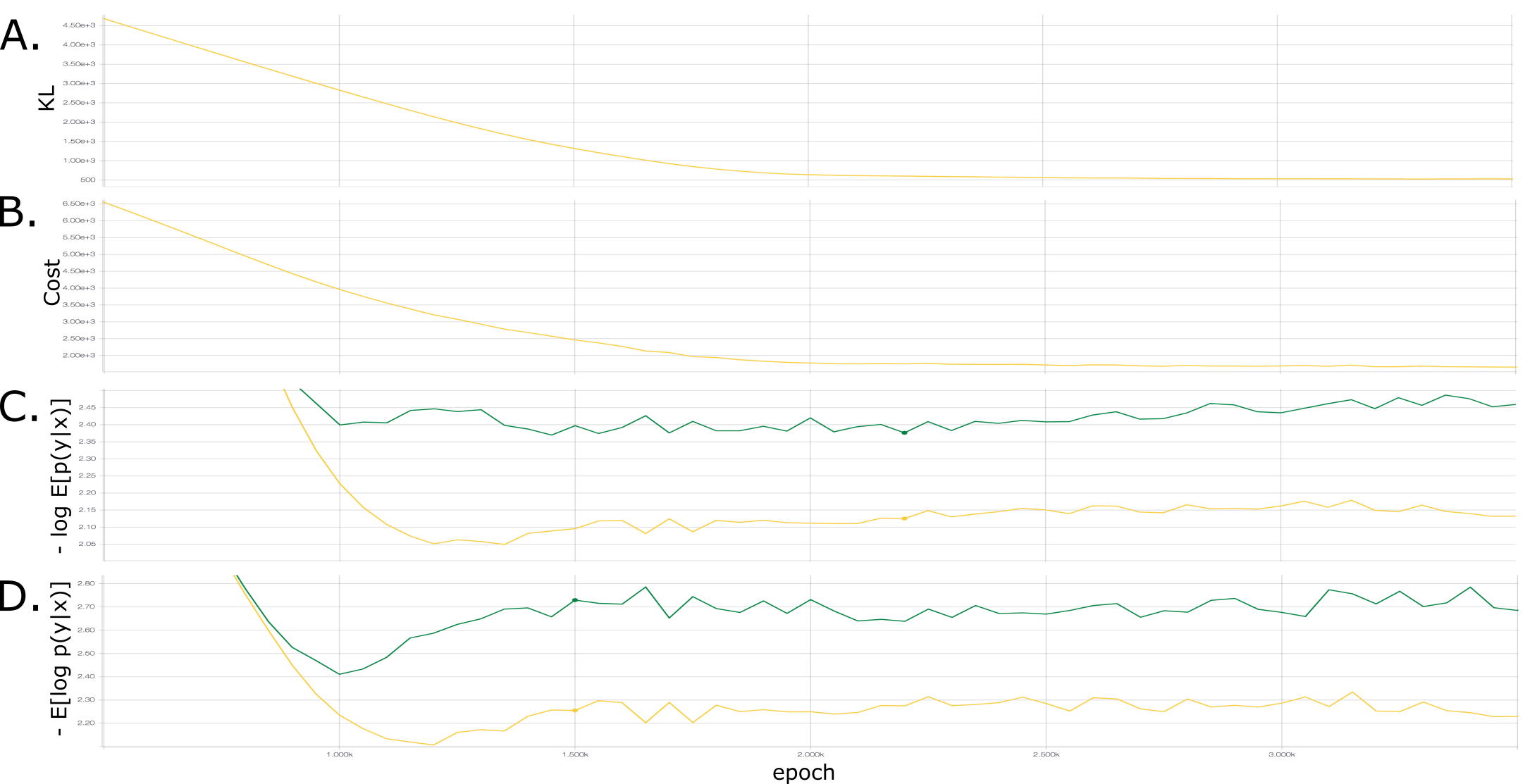} 
\caption{Optimization of variational Bayesian neural networks with mean field approximate posteriors.  Weight variances are initialized to be very small which leads to a large initial complexity penalty A.) the complexity penalty or `KL' term slowly decreases and converges.  B.) The overall variational free energy or `cost' slowly decreases and converges.  C.) the expected log likelihood and D.) The expected negative log likelihoods of both the test and training data under the posterior predictive distributions decrease initially but then increase again as the optimization proceeds and the model over-prunes.  In C and D, we additionally provide log likelihoods for the test set in green.}
\label{fig:pruning_curves}
\end{figure}

\begin{landscape}
\begin{figure}
\centering
\includegraphics[width=1.2\textwidth]{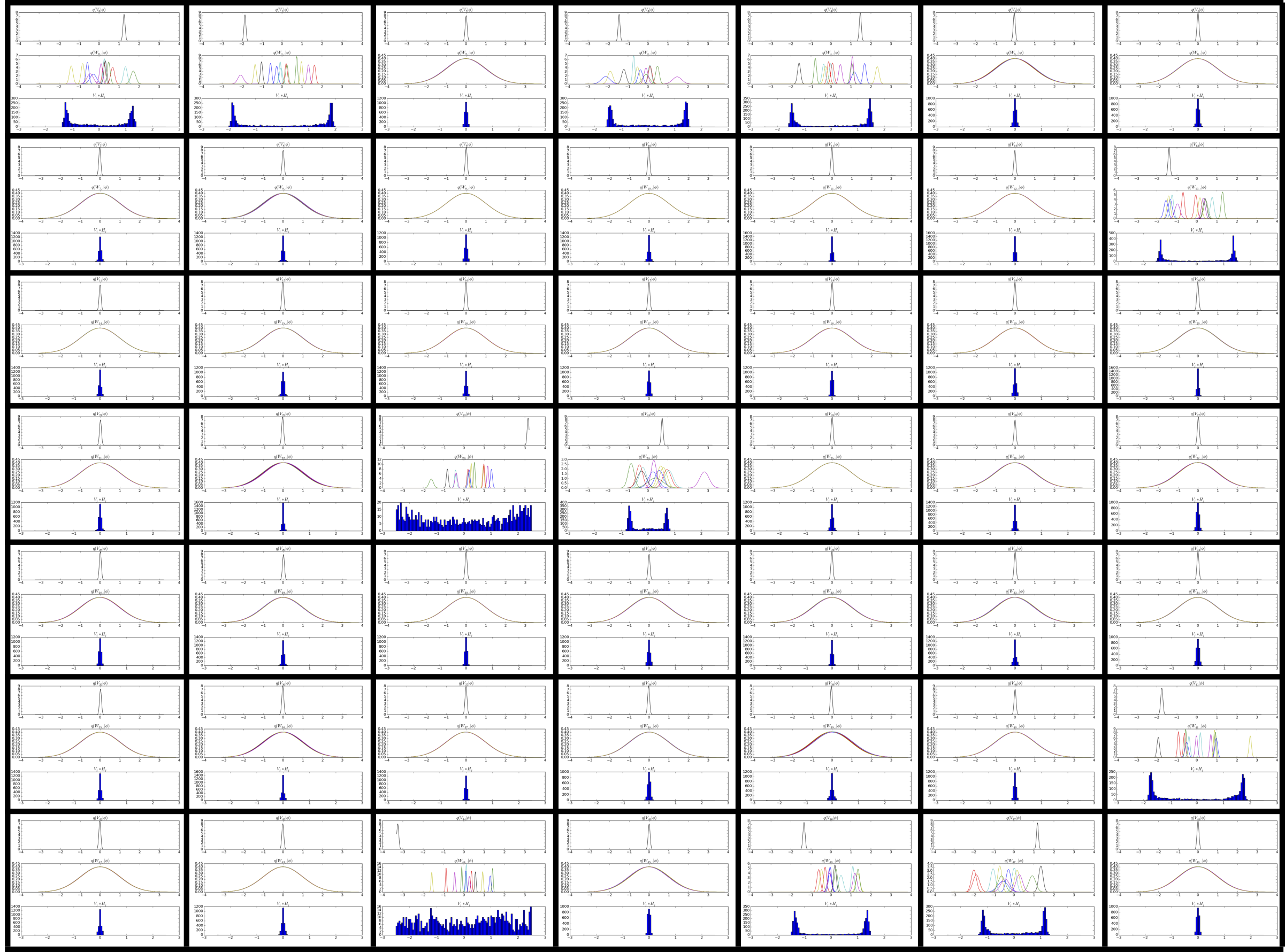}

\caption{Histograms depicting characteristics of each of the hidden units in of the neural network described in figure \ref{fig:pruning_rep}.  	 Top.) the approximate posterior over the weight connecting it to the output.  Middle.) posteriors over weights connecting from the input to the hidden unit Bottom.) Histogram of $25$ sampled activations across all training data-points.  All but $11$ of the hidden units have been pruned.  The pruned unit provided in the earlier figure (unit $45$) is not included. }
\label{fig:pruning_curves_App}
\end{figure}
\end{landscape}

\small
\end{document}